\documentclass[10pt, a4paper]{article}

\usepackage[final]{lrec2026} 

\usepackage{fontspec}

\usepackage{xcolor}
\usepackage{amsmath}
\usepackage{float}

\usepackage{booktabs}
\usepackage{multirow}
\usepackage{tikz} 
\usetikzlibrary{arrows.meta,positioning,shapes.multipart}

\newfontfamily\egyptology{NotoSerif-Regular.ttf}          
\newfontfamily\greek{NotoSerif-Regular.ttf}
\newfontfamily\coptic{NotoSansCoptic-Regular.ttf}         
\newfontfamily\ipa{NotoSerif-Regular.ttf}                 

\title{Semantic Alignment across Ancient Egyptian Language Stages via Normalization-Aware Multitask Learning
}

\name{He Huang} 

\address{Institut für Ägyptologie und Koptologie, Ludwig-Maximilians-Universität München \\
         Katharina-von-Bora-Straße 10, 80333 München, Germany \\
         h.huang@campus.lmu.de\\}

\abstract{
We study word-level semantic alignment across four historical stages of Ancient Egyptian. These stages differ in script and orthography, and parallel data are scarce. We jointly train a compact encoder-decoder model with a shared byte-level tokenizer on all four stages, combining masked language modeling (MLM), translation language modeling (TLM), sequence-to-sequence translation, and part-of-speech tagging under a task-aware loss with fixed weights and uncertainty-based scaling. To reduce surface divergence we add Latin transliteration and IPA reconstruction as auxiliary views. We integrate these views through KL-based consistency and through embedding-level fusion. We evaluate alignment quality using pairwise metrics, specifically ROC-AUC and triplet accuracy, on curated Egyptian–English and intra-Egyptian cognate datasets. Translation yields the strongest gains. IPA with KL consistency improves cross-branch alignment, while early fusion demonstrates limited efficacy. Although the overall alignment remains limited, the findings provide a reproducible baseline and practical guidance for modeling historical languages under real constraints. They also show how normalization and task design shape what counts as alignment in typologically distant settings.
 \\ \newline  }

\usepackage{fancyhdr}
\begin{document}

\maketitleabstract

\thispagestyle{fancy} 

\section{Introduction}

Ancient Egyptian remains understudied in natural language processing (NLP), despite preserving a rich set of etymological cognates through several historical stages. This paper focuses on aligning four representative stages grouped into two major branches: pre-Coptic Egyptian (Hieroglyphic and Demotic) and Coptic (Sahidic and Bohairic), the final stage of Ancient Egyptian. Although genealogically related, these stages differ significantly in script, vocabulary, and grammar. For the purpose of our cross-lingual alignment framework, we treat them as distinct languages and aim to align them within a unified model.

This task presents several challenges: (i) There are no closely related high-resource languages to support effective transfer learning. (ii) The scripts vary widely. Hieroglyphic and Demotic use Egyptological transliteration schemes with some special signs like {\egyptology ꜣ} and {\egyptology š}, while Coptic combines signs borrowed from Greek (e.g., {\coptic ⲁ}) with signs derived from Pre-Coptic Egyptian (e.g., {\coptic ϣ}). Only Coptic preserves vowels, while pre-Coptic Egyptian contains many homographs due to unwritten vowels, leading to semantic ambiguity. (iii) Egyptian texts often come from different thematic domains (e.g., indigenous Pre-Coptic Egyptian polytheistic versus Coptic Christian texts), and cross-stage parallel texts are rare, making direct alignment difficult.

We investigate how multitask learning can support word-level alignment under these constraints. Our setup includes masked language modeling (MLM), translation language modeling (TLM), sequence-to-sequence translation, and universal part-of-speech (UPOS) tagging, combined via dynamic loss weighting. To mitigate script inconsistency, we introduce normalization strategies (Latin-based and IPA) integrated through Kullback–Leibler (KL) regularization or early fusion. These also serve as data augmentation techniques for other low-resource languages facing similar transliteration issues. We evaluate alignment quality using AUC and triplet accuracy, which better reflect semantic similarity than top-$k$ recall in noisy, misaligned spaces. All code, data samples, and evaluation scripts are available in a GitHub repository.\footnote{\label{repo-footnote}Egyptian-alignment: \url{https://github.com/Merythuthor/Egyptian-alignment}. }

Rather than aiming for perfect alignment, our goal is to establish a realistic baseline and analyze the core difficulties. This work provides both methodological insights and empirical evidence for modeling historical language stages under minimal supervision.

\section{Related Work}

\paragraph{NLP for Egyptian and Coptic.}
Research on pre-Coptic Egyptian and Coptic has mainly focused on resource building and basic linguistic processing.  For a survey, see \citep{munoz-sanchez-2024-hieroglyphs}. Earlier works on Ancient Egyptian use resources from the project Thesaurus Linguae Aegyptiae (TLA) \citelanguageresource{TLA_Hiero, TLA_Demotic} and focuses on automated transliteration and automated translation from hieroglyphic signs or transliteration to English \citep{rosmorduc:hal-03118369, cao-etal-2024-deep, miyagawa-2025-rag}. For Coptic, the project Coptic SCRIPTORIUM \citelanguageresource{CopticScriptoriumResource} serves as a foundational ecosystem for digital studies \citep{SchroederZeldes2020Ecosystem}, providing annotated corpora, POS tagging, \citep{CopticPOSTagging2015} dependency treebanks \citep{ZeldesAbrams2018Treebank, zeldes-etal-2025-ud}, and core NLP pipelines \citep{ZeldesSchroeder2016Pipeline}. These resources have subsequently facilitated the development of low-resource language models, such as the development of the monolingual MicroBERT \citep{GesslerZeldes2022MicroBERT}.
While a few studies attempt to model multiple stages of Egyptian in a unified framework \citep{sahala-lincke-2024-neural, miyagawa-2025-rag}, semantic modeling across stages remains rare. Word-level semantic representation for Ancient Egyptian is largely understudied\citep{ORBi-2d0f5788-43d2-40c4-a02b-ec8a95030534}, which motivates our focus on word-level semantic alignment.

\paragraph{Alignment objectives and multitask learning.}
Multitask learning with objectives like MLM, TLM, and supervised translation promotes shared cross-lingual representations \citep{ConneauLample2019XLM,Ruder2019SurveyXling,Kendall2018MTLUncertainty}.  
Dynamic task weighting using uncertainty scaling has proven effective in balancing heterogeneous loss signals \citep{Kendall2018MTLUncertainty}. 

For ancient languages, research primarily targets Ancient Greek and Latin. Studies have compared post-hoc alignment of monolingual spaces versus joint multilingual training for Latin-Greek alignment \citep{wang2020crosslingualalignmentvsjoint}. Large language models have been evaluated for cross-lingual generalization across classical languages, demonstrating zero-shot transfer on various tasks \citep{akavarapu2025casestudycrosslingualzeroshot}. Regarding word-level alignment, multilingual models fine-tuned on Greek-Latin parallel data via successive training stages have achieved strong alignment accuracy \citep{yousef-etal-2022-automatic-translation}.

\paragraph{Normalization for historical texts.}
Orthographic normalization plays a key role in historical text processing. Prior work highlights that normalization can improve tagging and retrieval but may affect linguistic fidelity depending on the phonological depth and sparsity of the data \citep{Ruder2019SurveyXling,amrhein-sennrich-2020-romanization}. More recent work has demonstrated the benefits of transliteration for addressing script barriers in multilingual models \citep{liu-etal-2024-translico}. Consequently, our work treats normalization as an auxiliary representation and evaluates its effect on semantic alignment.

\section{Methods}\label{sec:method}
\subsection{Dataset} \label{subsec:dataset}
The datasets of this study come from two main sources. For the pre-Coptic Egyptian branch, we utilize Hieroglyphic and Demotic texts provided by the Thesaurus Linguae Aegyptiae (TLA) project \citelanguageresource{TLA_Hiero, TLA_Demotic}. The Coptic SCRIPTORIUM offers Sahidic and Bohairic texts from the Coptic branch \citelanguageresource{CopticScriptoriumResource}. Details regarding specific licensing constraints and reproducibility are provided in Section \ref{sec:data_code_availability}.

These four languages represent distinct stages of Ancient Egyptian, characterized by differences in script, orthography, and grammar. For illustrative examples of these diverse writing systems and their transliterated forms, see \hyperref[sec:appendix_a]{Appendix A}. Hieroglyphic encompasses multiple historical stages, but we merge them into one unit due to unclear internal boundaries and limited data. While Sahidic and Bohairic share many lexical items, they differ in spelling and morphosyntax. Therefore, we treat them as two languages in our experiments.

The language grouping into two branches informs both our model design and evaluation strategy. Alignment patterns are later analyzed at both intra- and cross-branch levels.

The original datasets include translations in English or German. Only entries with translations are selected for this study. German texts were translated into English using OPUS-MT (de$\rightarrow$en) \citep{tiedemann-thottingal-2020-opusmt}.
 UPOS tags are retained where available. Lemmatization is not used, as such annotations are missing for the Coptic data.

\subsection{Model}

We adopt a lightweight encoder-decoder architecture adapted to the four stages of Ancient Egyptian. Instead of fine-tuning a large multilingual model such as XLM-R, which is over-parameterized for our data and possesses a vocabulary incompatible with Egyptian transliteration, we train a shared, compact BERT encoder and a byte-level BPE tokenizer from scratch jointly on all four Egyptian stages. This design prevents out-of-vocabulary (OOV) fragmentation for special Unicode characters (e.g., {\egyptology ꜣ} and {\coptic ϣ}). English is incorporated in the encoder input through translation-related tasks, serving as a semantic pivot among different Egyptian languages.

The decoder is a standard Transformer initialized randomly and used only for sequence-to-sequence translation. The encoder and decoder share both the tokenizer and embedding matrix to enhance cross-representation consistency.

\begin{figure}[!ht]
\centering
\includegraphics[width=1.0\columnwidth]{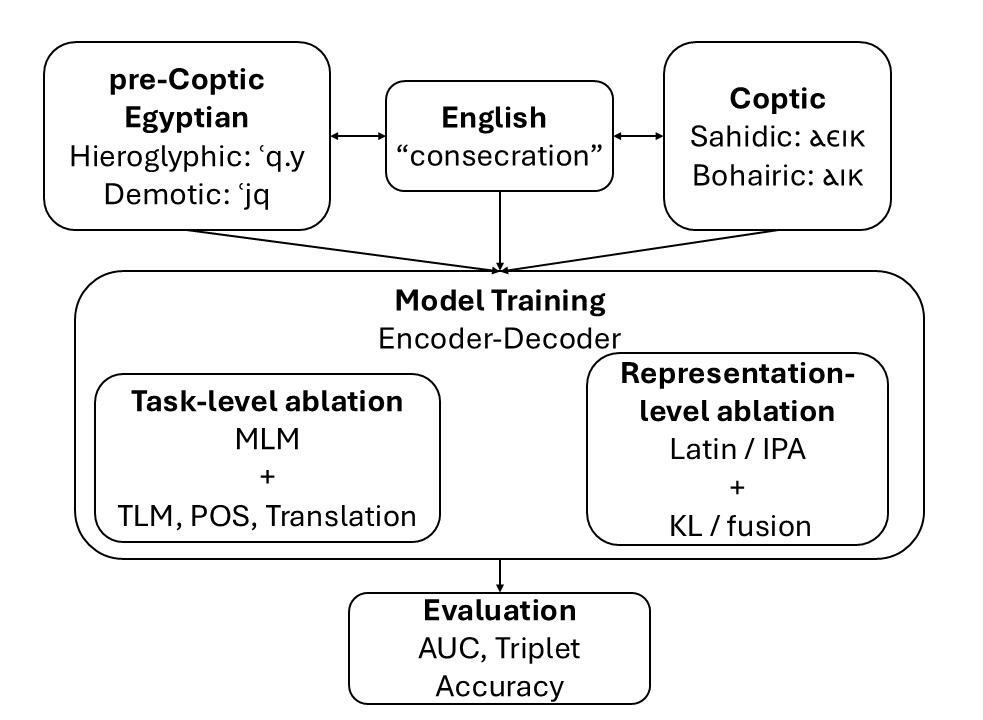}
\caption{Overview of the experimental pipeline. The ablation study isolates two independent dimensions: (1) task-level supervision, without normalization; and (2) representation-level inputs under a standalone MLM task.}
\label{fig:pipeline}
\end{figure}

Fig. \ref{fig:pipeline} summarizes our experimental workflow, including the four Egyptian language stages (illustrated by cognates for "consecration" in their respective orthographies), multitask setup, and evaluation process.

\subsection{Baseline and Task-Level Comparisons} \label{subsec:baseline}

As a baseline, we trained the model using MLM on the original texts from all four Egyptian languages. To help the model distinguish language identity despite using a shared tokenizer, each input sequence is prepended with a language-specific token (e.g., \verb|<hiero>|, \verb|<dem>|, \verb|<sah>|, \verb|<boh>|).

We implement a multitask framework combining MLM, TLM, translation and UPOS tagging. While MLM learns token-level representations independently per language, the cross-lingual tasks (TLM, translation) use English (which is prepended with the \texttt{<eng>} token) to provide alignment signals. TLM jointly masks the Egyptian and English sequences, encouraging attention-based token alignment across the encoder (see \hyperref[sec:appendix_b]{Appendix B}). 

\smallskip
\noindent\textbf{Example of TLM input:} \texttt{[CLS] <hiero> nfr sw [SEP] <eng> he is good [SEP]}.
\smallskip

Sequence-to-sequence translation leverages decoder cross-attention to directly guide alignment. UPOS tagging is used as an auxiliary task to provide syntactic supervision. We do not apply dictionary-based contrastive supervision due to noise and lack of lemma information in Coptic.

To balance the influence of each task, we apply a hybrid loss combining static weights and uncertainty-based adaptive scaling \citep{Kendall2018MTLUncertainty}. For task \(i\), the total loss is:

\begin{equation}
\mathcal{L}_{\text{total}} = \sum_i W_i \left( \frac{1}{\sigma_i^2} \mathcal{L}_i + \log \sigma_i^2 \right),
\end{equation}

where \(W_i\) is a fixed prior weight and \(\sigma_i\) is a learnable parameter capturing task uncertainty. This design improves robustness in multitask training while allowing dynamic adjustment based on learning difficulty. Further implementation details and ablation settings are described in Section~\ref{sec:experiment}.

\subsection{Representation-Level Normalization Strategies}\label{subsec:normalization}

To mitigate surface divergence caused by the heterogeneous character sets across Egyptological transliteration and Coptic script, we apply two normalization methods: Latin transliteration and IPA-based phonemic reconstruction. These are used as auxiliary views rather than replacements, with the original form preserved throughout training.

Unlike Egyptological transliteration with its special phonograms, the Latin normalization maps all graphemes to standard Latin characters (e.g., {\egyptology š}, {\coptic ϣ} → sh) (see sample text in \hyperref[sec:appendix_b]{Appendix B}). Though not fully phonologically precise, it increases token overlap and highlights etymological similarities. The IPA scheme uses approximate phonemic representations to restore missing vowels in the pre-Coptic stages and unify consonant encoding. For instance, we render weak consonants like {\egyptology ꜣ}, {\egyptology ꜥ}, {\egyptology ꞽ}, and w between two consonants as vowels \textit{aa}, \textit{a}, \textit{i} and \textit{u}, and fill remaining consonant gaps with \textit{e}. This transforms \textit{nfr} (Hieroglyphic, “good”) into \textit{nefer}, aligning it more closely with its Coptic cognate {\coptic ⲛⲟⲩϥⲉ} “good”, converted into \textit{noufe}.

Full normalization mappings are provided as \texttt{training/normalization.py} in the GitHub repository (see footnote~\ref{repo-footnote}). English text is not normalized into IPA to prevent misleading the model with coincidental sound overlap across unrelated language families.

We acknowledge that such normalization introduces simplifications and potential semantic drift, especially given the complexity of Egyptian phonology and its transliteration systems. Nonetheless, this rudimentary attempt serves as a first step toward understanding how orthographic normalization may aid cross-lingual semantic modeling, offering valuable guidance for future refinement.

\subsection{View Integration Methods}

To preserve the original orthography to retain semantic richness while providing phonological regularity, we compare two methods for integrating these auxiliary views into training:

\paragraph{(A) Consistency via  KL Divergence.}  

We adopt symmetric Kullback–Leibler (KL) divergence to enforce bidirectional alignment between the predictive distributions of original and normalized forms, effectively performing mutual distillation across orthographic views.

Given the token-level MLM predictions from both views, we minimize:
\begin{multline}
\mathcal{L}_{\text{consistency}} =
\frac{1}{2} \Big[ D_{\mathrm{KL}}(p_{\text{orig}} \parallel p_{\text{norm}}) \\
+ D_{\mathrm{KL}}(p_{\text{norm}} \parallel p_{\text{orig}}) \Big],
\end{multline}

where gradients are detached on the teacher side to prevent representational collapse.

The total loss is:

\begin{multline}
\mathcal{L}_{\text{total}} =
W_{\text{MLM}} \left(
\frac{1}{\sigma_{\text{MLM}}^2} \mathcal{L}_{\text{MLM}}
+ \log \sigma_{\text{MLM}}^2
\right) \\
+ \lambda_{\text{KL}} \cdot \mathcal{L}_{\text{consistency}},
\end{multline}

with \(\lambda_{\text{KL}}\) as a fixed hyperparameter (\texttt{--consistency\_lambda 0.5}).

This regularization encourages the encoder to learn view-invariant representations across orthographic variants. While this strategy is applicable to tasks beyond MLM (e.g., translation or POS tagging), our experiments restrict it to MLM due to computational resource limits.

\paragraph{(B) Early Fusion via Embedding Mixture.}  

We integrate the original and normalized views at the embedding level by constructing a learnable weighted mixture, enabling soft alignment without requiring multiple forward passes.

Given token embeddings \(E_{\text{orig}}\) and \(E_{\text{norm}}\), we compute a fused representation:
\begin{equation}
E_{\text{fused}} = \alpha \cdot E_{\text{orig}} + (1 - \alpha) \cdot E_{\text{norm}},
\end{equation}
where \(\alpha \in (0, 1)\) is derived from a learnable scalar via sigmoid activation.

The fused embedding is passed to the encoder to compute the MLM loss as usual. The total loss is:
\begin{equation}
\mathcal{L}_{\text{total}} =
W_{\text{MLM}} \left(
\frac{1}{\sigma_{\text{MLM}}^2} \mathcal{L}_{\text{MLM}} + \log \sigma_{\text{MLM}}^2
\right).
\end{equation}

Unlike KL divergence, this method requires only one forward pass per example and introduces no auxiliary loss term, making it computationally efficient. The model learns to balance both views during training:
when \(\alpha \to 1.0\), it favors the original form; when \(\alpha \to 0.0\), it prefers the normalized variant. This soft fusion lets the model learn representations that incorporate phonological regularity while retaining language-specific features under a single, global learned weighting parameter.

\paragraph{Strategy Comparison.}

These strategies address complementary aspects of multi-representation learning. KL consistency provides output-level regularization via bidirectional distillation, whereas early fusion performs input-level soft integration. Together, they allow us to empirically evaluate whether early fusion outperforms late-stage consistency in capturing script-invariant semantics under low-resource constraints. 

It is important to note an implementation detail regarding both strategies: our current formulations of KL consistency and early fusion operate via strict position-wise alignment across padded subword sequences. We analyze the structural effects of this design choice in Section \ref{sec:error_analysis}.

\section{Experiments}\label{sec:experiment}
\subsection{Dataset Preparation} 

We standardized lacuna markers (e.g., `---`, `...`, `<gap>`) into one or more \texttt{[gap]} tokens. Sentences with gaps were kept. To retain essential morphological and syntactic information, we preserved key linguistic markers, which include the period (\texttt{.}), separating stems from gender and number endings, and the suffix sign ({\egyptology ⸗}) in pre-Coptic Egyptian, and the hyphen (\texttt{-}).

Table~\ref{tab:corpus-stats} shows corpus statistics after preparation. To preserve the natural distribution of the attested corpora, we did not apply any sentence-level resampling during tokenizer or model training to mitigate the imbalance across Egyptian languages.  

\begin{table}[!ht]
\centering
\small

\makebox[\columnwidth][c]{%
\begin{tabular}{@{}|l|r|r|r|r|@{}}  
\hline
\textbf{Lang} & \textbf{Sentences} & \textbf{Sent\%} & \textbf{Tokens} & \textbf{Tok\%} \\
\hline
H & 145{,}060 & 65\% & 2{,}705{,}136 & 56\% \\
D & 30{,}488  & 14\% &   646{,}327   & 13\% \\
S & 25{,}902  & 12\% &   800{,}198   & 17\% \\
B & 20{,}538  &  9\% &   693{,}097   & 14\% \\
\hline
\textbf{TOTAL} & \textbf{221{,}988} & \textbf{100\%} & \textbf{4{,}844{,}758} & \textbf{100\%} \\
\hline
\end{tabular}%
}
\caption{Corpus size and distribution. Lang = language stage (H = Hieroglyphic, D = Demotic, S = Sahidic, B = Bohairic). Token counts are computed using the shared byte-level BPE tokenizer (excluding special tokens).}

\label{tab:corpus-stats}
\end{table}

Each entry consists of the original transcription, its English translation, and a language identifier. No normalization is applied during preprocessing; instead, normalization is introduced during training, as described in Section~\ref{subsec:normalization}.

\subsection{Training Setup}
\label{subsec:training_setup}
We trained a shared byte-level BPE tokenizer (vocab size: 32{,}000, min frequency: 2) on the combined corpus of all four Egyptian language stages and their English translations. The vocabulary also includes task-specific tokens (e.g., \texttt{[CLS]}, \texttt{[MASK]}), language tags (e.g., \texttt{<hiero>}), and the gap marker \texttt{[gap]}. To isolate the effect of cross-lingual alignment mechanisms, English data were strictly limited to in-domain parallel texts. 

We intentionally excluded large-scale monolingual English corpora for three main reasons: (i) introducing massive English data would lead to severe over-representation, causing it to disproportionately dominate the shared vocabulary and model capacity; (ii) modern English corpora introduce significant domain shift, as contemporary vocabulary diverges from the specific historical and religious contexts of Ancient Egyptian, thereby injecting noise into the shared semantic space; and (iii) under our computational constraints, a restricted, in-domain setup maximizes training efficiency without wasting resources on irrelevant data.

The tokenizer was trained exclusively on the original historical scripts and the English translations, without exposure to the normalized Latin and IPA views. While this strictly grounds the vocabulary in the attested corpora, it forces the tokenizer to over-segment the unseen normalized inputs into much shorter subwords (borrowed from English) or raw character fallbacks, leading to higher sequence fragmentation.

The dataset is split into training/validation/test sets in an 8:1:1 ratio. 

All models are trained for 10 epochs using the encoder-decoder architecture described in Section~\ref{sec:method}, with 6 encoder and 6 decoder layers, hidden size 768, 12 attention heads, and max sequence length 768. The encoder and decoder share embeddings. Training was performed on a single NVIDIA RTX 5090 GPU with a batch size of 16, learning rate of $5\cdot10^{-5}$, bfloat16 mixed precision, cosine decay with 500 warm-up steps, and gradient accumulation set to 2.

\subsection{Ablation Design}

We perform ablation studies along two independent dimensions, using an MLM-only model trained on the original texts (without normalization) as our unified baseline.

\paragraph{Task-level ablation.}
To evaluate the contribution of each auxiliary task, we selectively activate or deactivate them during training with the original scripts, without normalization. When a task is active, its fixed prior weight ($W_i$, as defined in Section~\ref{subsec:baseline}) is set to a predefined value ($W_{\text{MLM}}=1.0$, $W_{\text{Trans}}=1.0$, $W_{\text{TLM}}=1.0$, and $W_{\text{POS}}=0.5$). To ablate a task, its weight is set to zero. The MLM objective is always retained ($W_{\text{MLM}}=1.0$).

\paragraph{Representation-level ablation.}
To assess the impact of normalization, we evaluate three input formats under the MLM-only setup: (1) raw orthography, (2) Latin transliteration, and (3) IPA-based phonemic representation. Each normalization is paired with either KL consistency (fixed $\lambda_{\text{KL}}=0.5$)  or early fusion (trainable $\alpha$, initialized to 0.5) to evaluate integration strategies.

\section{Evaluation}

Our evaluation focuses on cross-lingual semantic alignment at the word level across four historical stages of Ancient Egyptian. As our objective is not sequence generation or syntactic tagging, we omit BLEU and POS F1 as primary metrics.

\subsection{Visualization of Clustering Geometry} 
We begin by visualizing the structure of the multilingual embedding space. This subsection illustrates why nearest-neighbor search may be unreliable in our setting. Instead of relying on top-$k$ lexical retrieval, which assumes a well-aligned semantic space, we use relative distance-based metrics (see Subsection~\ref{subsec:metrics}). 

For each word, we compute its embedding by mean-pooling subword vectors over all occurrences in the validation and test sets. We visualize 500 (or 400, when including English) word-level embeddings per language, sampled with a frequency of $\geq$3, using t-SNE (perplexity=30, init=PCA, n\_iter=2000) after L2 normalization. All embeddings are produced using the shared byte-level BPE tokenizer.

\begin{figure}[!ht]
\begin{center}
\includegraphics[width=\columnwidth]{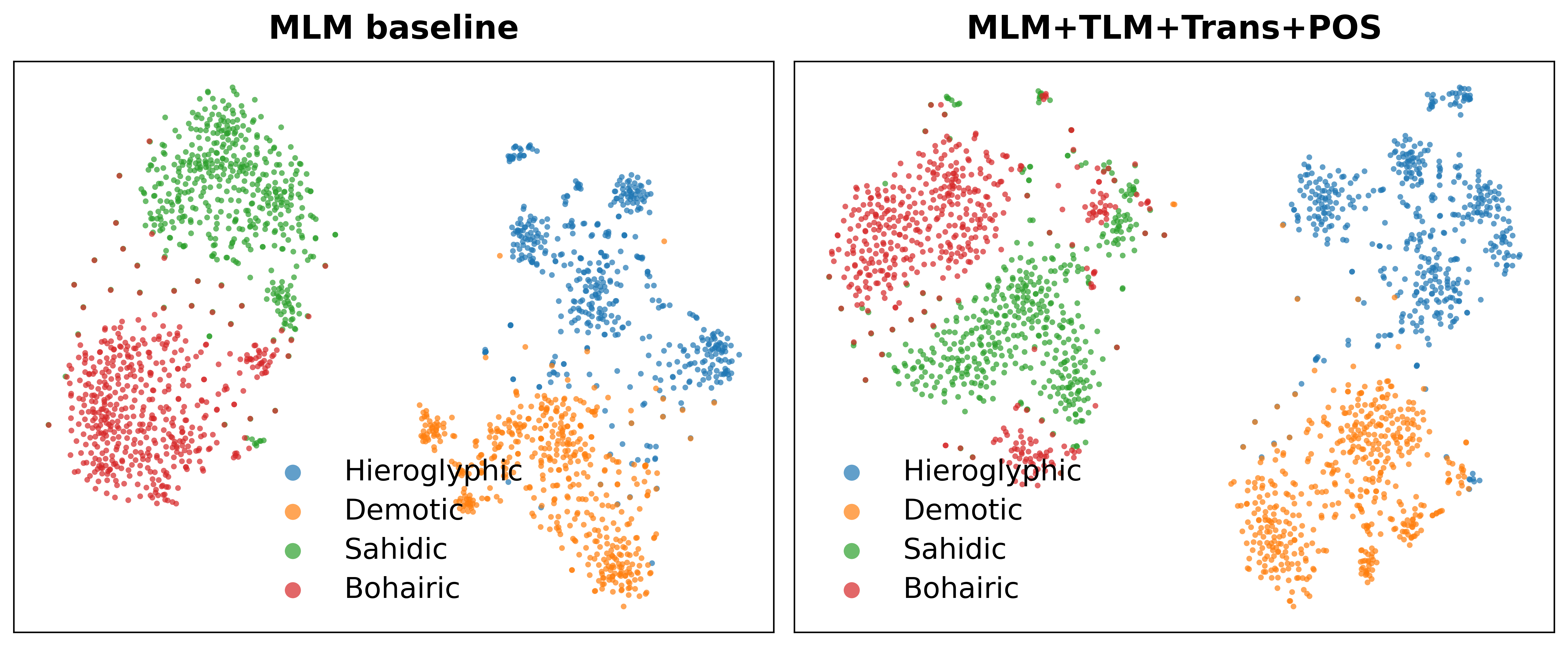}
\caption{t-SNE visualization of word embeddings across four Egyptian language stages through the MLM-only baseline and the full multitask model (MLM+TLM+Translation+POS).}
\label{fig:tsne-egypt}
\end{center}
\end{figure}

Fig.~\ref{fig:tsne-egypt} shows the embedding space for the four Egyptian languages. The MLM-only baseline yields four distinct language clusters. The languages with their respective branches naturally cluster together. Since internal coherence is a prerequisite for alignment, the left panel suggests that while auxiliary supervision in multitask training (MLM+TLM+Translation+POS) might improve alignment within the Coptic branch, it weakens the coherence of pre-Coptic Egyptian scripts. Consequently, overall alignment remains limited. Most word embeddings stay concentrated within their respective language clusters, with minimal cross-lingual overlap.

\begin{figure}[!ht]
\begin{center}
\includegraphics[width=\columnwidth]{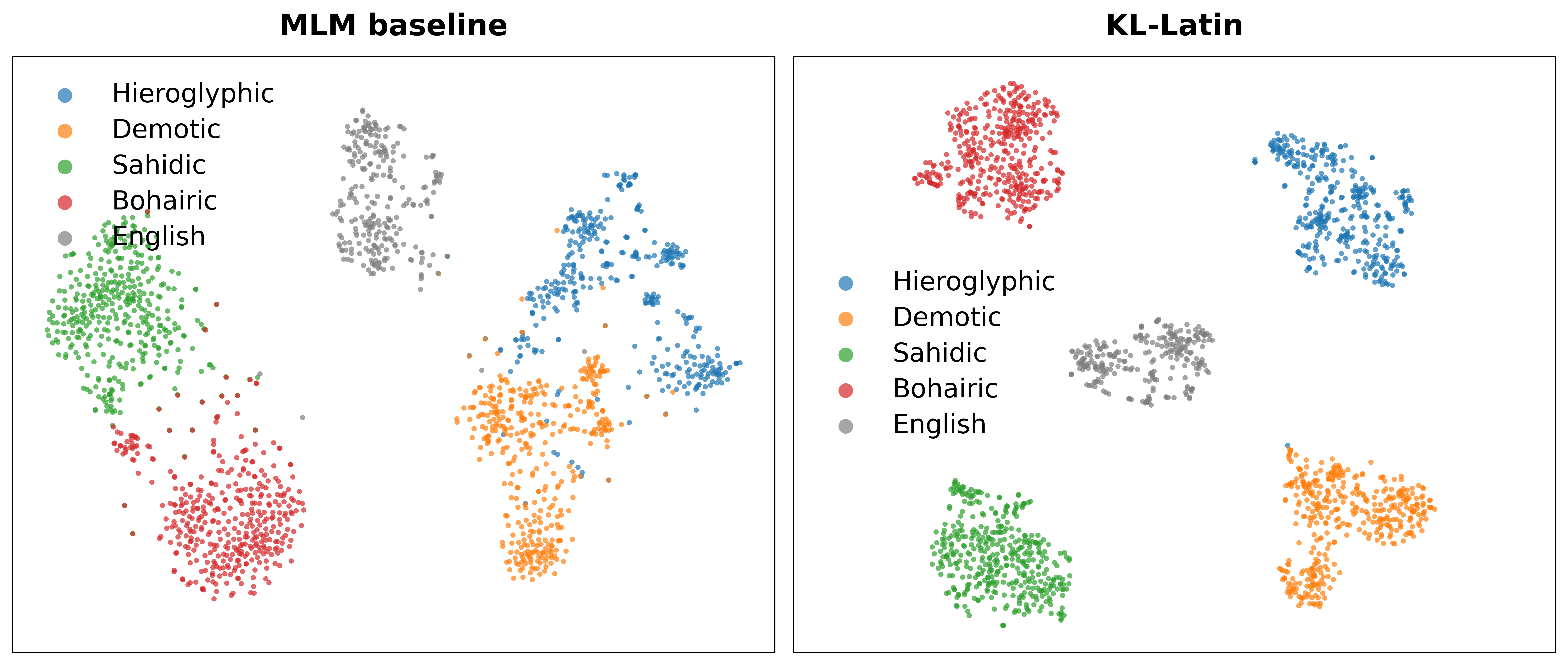}
\caption{t-SNE visualization including English as a pivot language through the MLM-only baseline with or without Latin normalization and KL consistency training.}
\label{fig:tsne-egypt-english}
\end{center}
\end{figure}

Fig.~\ref{fig:tsne-egypt-english} includes English as a fifth language through the MLM-only baseline. English acts as a semantic pivot, inducing radial clustering and hubness. While this improves global comparability, it distorts local neighborhoods among Egyptian variants. Incorporating Latin normalization via KL consistency increases the global spatial separation of the Egyptian clusters instead of merging them. We attribute this paradox to normalization acting as a strong regularizer that strips away surface noise while harmonizing surface orthography. Consequently, the MLM objective is forced to cluster tokens by underlying language-specific morphosyntactic structures, isolating each language into a distinct subspace.

Nevertheless, it is important to note that t-SNE visualizations primarily illustrate macro-level topological clustering and do not strictly quantify word-level semantic alignment precision. Because the embeddings are globally segregated by language, direct top-$k$ retrieval fails to provide a faithful measure of cross-lingual semantic overlap under current settings and should only be used after successful normalization or alignment strategies are applied.

\subsection{Evaluation Setup and Datasets} 
    
We assess semantic alignment using alignment-agnostic metrics, namely triplet accuracy and ROC-AUC, computed over curated positive and negative word pairs. These metrics assume only that true cognates should be closer than unrelated pairs on average. Word embeddings are computed by averaging subword vectors from the shared BPE tokenizer.
    
This setup is robust to global embedding shifts and script-based offsets, making it suitable for low-resource, cross-script scenarios. Unlike top-$k$ retrieval, which assumes the existence of well-aligned vector spaces, these relative distance-based metrics evaluate relative semantic proximity without requiring absolute alignment. Due to the genre divergence across stages (e.g., indigenous pre-Coptic Egyptian religious vs. Coptic biblical texts), shared contexts are rare. We thus compute word embeddings by averaging over all occurrences and fix the random seed for consistency.

In this framework, we evaluate semantic alignment at two levels:
    
\paragraph{Egyptian–English.}
We collect dictionary pairs from TLA and Coptic SCRIPTORIUM. These test whether the model can associate Egyptian words with their corresponding English translations. To assess generalization, we split the paired examples into \textit{Seen} (where the Egyptian word and its English equivalent co-occurred within the same sentence-translation pair during training) and \textit{Unseen}. A strong model should perform well even on \textit{Unseen} pairs.
    
\paragraph{Intra-Egyptian.}
The intra-Egyptian evaluation dataset is based on cognate pairs among Egyptian languages produced by TLA, which are accessed under research terms and cannot be redistributed. We remove duplicates and low-frequency terms and distinguish between Sahidic and Bohairic forms. To illustrate the data structure, we provide sample pairs of these cognate pairs in \hyperref[sec:appendix_c]{Appendix C}, and include ten example groups in \texttt{/resource\_eval/} in the GitHub repository (see footnote~\ref{repo-footnote}).

We categorize examples as:
\begin{itemize}
    \item \textbf{Cross-branch:} e.g., Hieroglyphic–Sahidic
    \item \textbf{Within-branch:} e.g., Sahidic–Bohairic
    \item \textbf{Homographs vs. Heterographs:} Defined by whether the forms share identical spelling
\end{itemize}

This evaluation directly addresses our core research question: can models capture semantic correspondence across orthographically divergent but genealogically related stages?

\subsection{Evaluation Metrics}\label{subsec:metrics}

All metrics operate at the word level, using cosine similarity \(\text{sim}(\cdot, \cdot)\). For each language pair, we use up to \(N\) gold pairs \(\mathcal{P} = \{(w_i, w_i^+)\}_{i=1}^{N}\) (detailed evaluation dataset statistics and valid pair counts are provided in \hyperref[sec:appendix_d]{Appendix D}), and construct matched negatives \(\mathcal{N} = \{(w_i, w_i^-)\}_{i=1}^{N}\) by randomly sampling non-cognates from the same target language.

\paragraph{Triplet Accuracy.}
Measures whether each true pair is more similar than its negative counterpart:
\begin{equation}
\text{Acc} = \frac{1}{N} \sum_{i=1}^{N} \mathbf{1}\left[ \mathrm{sim}(w_i, w_i^+) > \mathrm{sim}(w_i, w_i^-) \right]
\end{equation}

The result reflects the proportion of triplets where the model assigns higher similarity to the correct pair. Since each comparison is local and independent, this metric is sensitive to the model’s decision for each word. The expected value of a random model is 50\%.

\paragraph{ROC-AUC.}
AUC is computed over the entire distributions of positive and negative similarity scores. It estimates the probability that a randomly chosen positive pair is more similar than a randomly chosen negative:

\begin{equation}
\text{AUC} = \frac{1}{|\mathcal{P}||\mathcal{N}|} \sum_{p \in \mathcal{P}} \sum_{n \in \mathcal{N}} \mathbf{1}\big[ \mathrm{sim}(p) > \mathrm{sim}(n) \big]
\end{equation}

This metric considers all cross-pair combinations between positive and negative sets. It is invariant to monotonic transformations and global shifts in embedding space, providing a holistic view of score separability.

Together, Triplet Accuracy and ROC-AUC offer a comprehensive view of alignment performance in noisy, cross-script scenarios where top-\(k\) recall is unreliable.

\section{Results}

\begin{figure}[!ht]
\begin{center}
\includegraphics[width=\columnwidth]{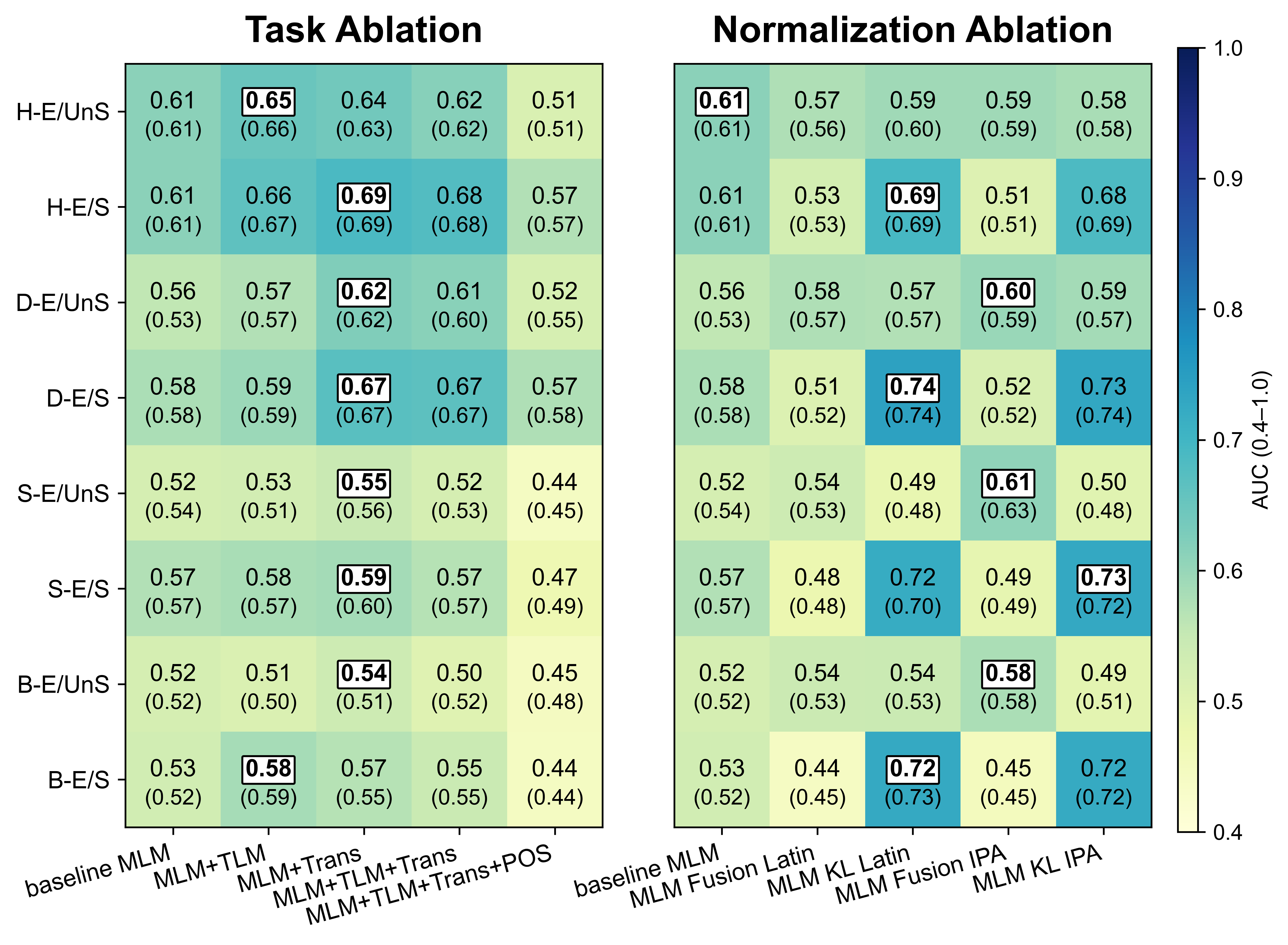}
\caption{Alignment performance (AUC and Accuracy) between Ancient Egyptian and English under different multitask and normalization settings. Each cell displays AUC (top) and Accuracy (bottom, in parentheses), rounded to two decimal places. The colormap encodes AUC scores from 0.40 to 1.00, which serves as the primary metric; Accuracy is shown for reference. The highest AUC in each row is highlighted with a white rectangle.}
\label{heatmap-egyptian-english}
\end{center}
\end{figure}

\begin{figure}[!ht]
\begin{center}
\includegraphics[width=\columnwidth]{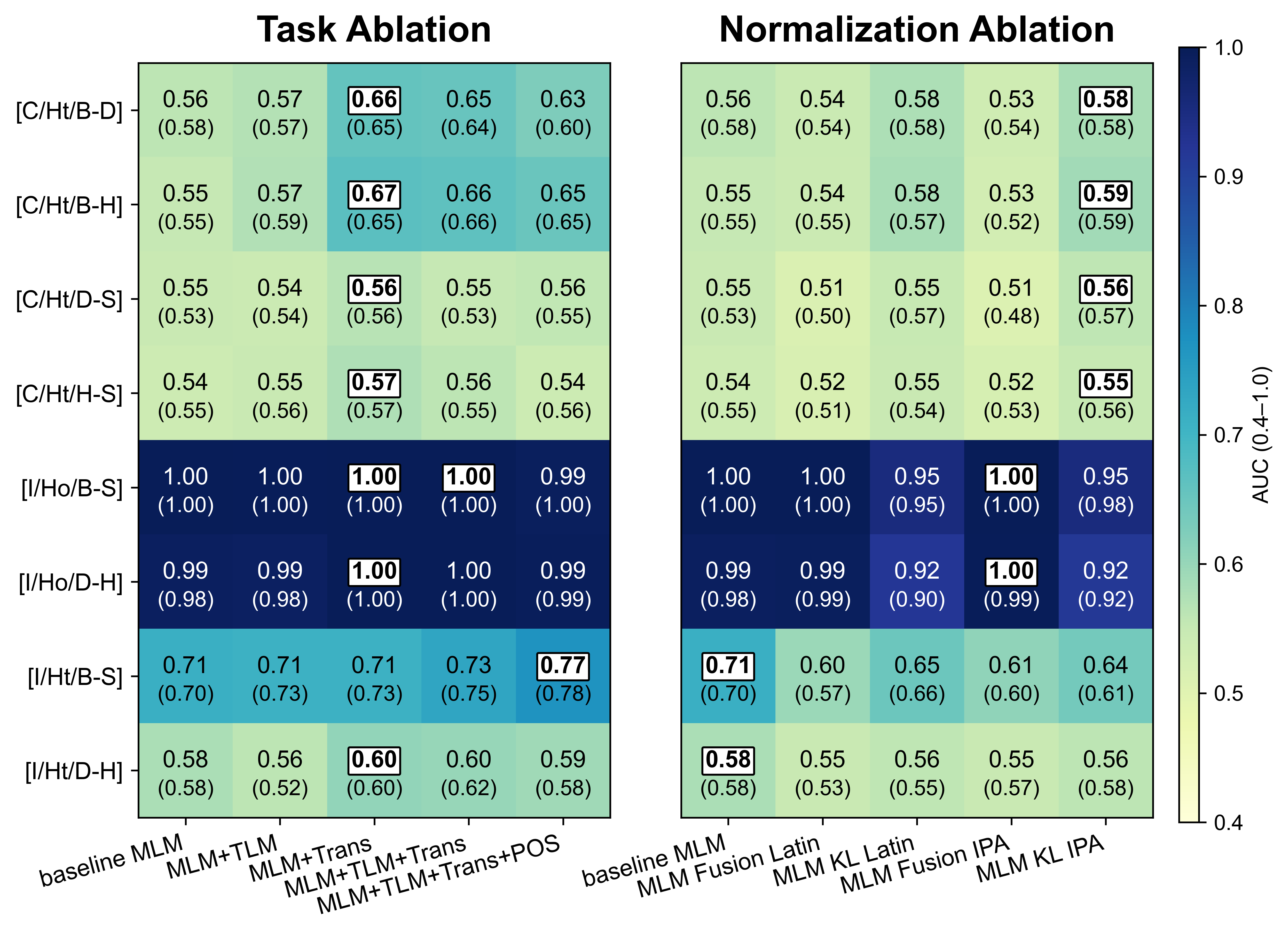}
\caption{Intra-Egyptian alignment performance (AUC and Accuracy) across historical stages and training variants. The highest AUC in each row is highlighted with a white rectangle.}
\label{heatmap-egyptian}
\end{center}
\end{figure}

Our main results are presented in two heatmaps detailing alignment performance across tasks and normalization strategies (Fig.~\ref{heatmap-egyptian-english} and Fig.~\ref{heatmap-egyptian}). They report Egyptian–English results and intra-Egyptian alignment respectively, and will be analyzed separately in the following subsections.

In both figures, the left subfigure varies task settings (MLM, +TLM, +Translation, +POS), while the right subfigure varies representation strategies (Latin, IPA) and integration methods (KL consistency, early fusion). The highest AUC in each row is highlighted with a white rectangle.

\textbf{Notation:} C = cross-branch; I = intra-branch; Ht = heterograph; Ho = homograph; D/H/S/B = Demotic/Hieroglyphic/Sahidic/Bohairic; E = English; S/U = \textit{Seen/Unseen}.

\subsection{Result Analysis: Egyptian-English}
\paragraph{Task Ablation}
As shown in the left panel of Fig.~\ref{heatmap-egyptian-english}, AUC and Accuracy follow consistent trends. \textit{Seen} pairs slightly outperform \textit{Unseen} pairs, suggesting that training provides moderate alignment gains. Among Egyptian varieties, Demotic and Hieroglyphic achieve better alignment with English than the Coptic languages. While Hieroglyphic benefits from a larger corpus, Demotic's performance is notable despite its smaller size. This may stem from their shared orthographic overlap with English, with the exception of a few special signs. In contrast, Coptic uses Greek-based characters, making surface alignment with English harder, a pattern further confirmed in the normalization experiments.

Across training objectives, two-task models generally outperform MLM-only. MLM+Translation consistently yields stronger alignment than MLM+TLM. Adding POS offers no clear improvement over two-task setups despite being a word-level task, highlighting the need for careful multitask design.

\paragraph{Normalization Ablation}

On the right panel of Fig.~\ref{heatmap-egyptian-english}, early fusion performs poorly. In most cases, the \textit{Seen} scores are even lower than \textit{Unseen}, and the overall performance is similar to or below the MLM baseline. This indicates that early fusion tends to disrupt rather than enhance cross-lingual alignment. Nevertheless, when combined with IPA normalization, early fusion outperforms the other groups in the \textit{Unseen} scenario for Demotic, Sahidic, and Bohairic. This unexpected pattern warrants further investigation in future work.

In contrast, KL-based consistency training yields a clear \textit{Seen-Unseen} gap in favor of \textit{Seen}, indicating that the model captures stronger alignment signals in this setting. Both IPA and Latin normalization show comparable results on \textit{Seen} pairs and perform much better than the model without normalization. On \textit{Unseen} pairs, the improvement is smaller, suggesting limited generalization ability. 

Across the four stages, Demotic, Sahidic, and Bohairic align more closely to English than Hieroglyphic. This outcome is not fully explained by corpus size alone, since Hieroglyphic is also the largest corpus. Domain differences across sources may be a factor, but require further analysis. Overall, normalization under KL consistency improves Egyptian–English alignment, while early fusion remains ineffective and sometimes counterproductive.

\subsection{Result Analysis: Intra-Egyptian}
\paragraph{Task Ablation}

AUC and Accuracy in Fig.~\ref{heatmap-egyptian} show similar trends. Alignment is stronger within branches than across, likely due to shared transliteration and more identical cognate forms. In cross-branch cases, adding translation helps Bohairic align better to the pre-Coptic stages than Sahidic, despite having less data. This is likely due to Bohairic's limited corpus size and restricted semantic domain, allowing semantic alignment via English to be more easily learned.

Homographs within branches reach near-perfect alignment. Even the MLM-only baseline achieves strong results without cross-lingual supervision. This within-branch alignment is primarily driven by their shared transliteration schemes. As illustrated in \hyperref[sec:appendix_a]{Appendices A} and \hyperref[sec:appendix_c]{C}, Hieroglyphic and Demotic pairs are often transcribed into identical or highly similar surface forms.

For heterographs, Sahidic–Bohairic performs markedly better than Demotic–Hieroglyphic. This indicates that the Coptic branch exhibits stronger orthographic distinctiveness. Because Coptic preserves vowels, words with different meanings are more clearly differentiated from each other, allowing true cognates to remain recoverable even with minor dialectal spelling variations.

Pre-Coptic Egyptian, however, contains a much greater number of homographs due to unwritten vowels. In the original script, semantic ambiguity was resolved by iconographic features like determinatives, but these crucial visual cues are lost in the transliterations used for training. Consequently, even identical phonetic representations of consonants can map to multiple distinct meanings, severely complicating the disambiguation of homographs and the alignment of heterographs. 

This observation strongly motivates adding lemma identifiers where available to disambiguate pre-Coptic Egyptian homographs, while also suggesting that the absence of lemma IDs on the Coptic side may be less damaging than initially expected.

Across multitask settings, MLM+Translation consistently outperforms other settings. Adding TLM or POS to this setup brings little further gain. This implies that, under limited word-level supervision, sequence-level translation signals are the strongest driver of intra-Egyptian alignment. 

Unlike in Egyptian–English alignment, POS does not harm performance here, likely due to shared underlying syntax across Ancient Egyptian stages. Adding POS especially helps the alignment between dialectal variants of Sahidic and Bohairic. This is probably due to the syntactic stabilization during the Coptic stage, when grammar became more standardized and structured than in the pre-Coptic periods.

\paragraph{Normalization Ablation}
The vertical patterns across evaluation groups largely match the task ablation trends and are not repeated here. Focusing on normalization choices, IPA with KL consistency gives the best cross-branch alignment. A likely reason is that IPA partially restores the vowel structure of pre-Coptic Egyptian, facilitating connections to Coptic varieties containing vowels. The gains, however, are modest, indicating that finer-grained phonological reconstruction rules are needed.

In contrast, for within-branch heterographs, normalization often reduces alignment quality. This supports the view that normalization can remove useful language-specific cues and distort originally informative spellings. When surface forms are already similar, further normalization may harm both semantic proximity and other latent signals. These findings support adopting selective normalization policies. Phonology-oriented normalization should be applied only for distant scripts and cross-branch comparisons.

\subsection{Error Analysis: The Subword Fragmentation Bottleneck}
\label{sec:error_analysis}

While translation yields strong gains, early fusion and KL consistency performed below expectations in certain configurations (e.g., early fusion consistently underperforming the baseline). To understand this, we conducted a qualitative error analysis on the tokenized inputs (see \hyperref[sec:appendix_b]{Appendix B}).

We identified a structural bottleneck rooted in subword-level positional misalignment. As noted in Section~\ref{subsec:training_setup}, the shared BPE tokenizer was not exposed to the normalized views during training. Consequently, Latin and IPA sequences suffer from severe over-segmentation. For example, a single word in the original script might be tokenized into two subwords, while its Latin or IPA equivalent is fragmented into four or more short subwords or fallback tokens.

Because our integration mechanisms (KL consistency and early fusion) currently rely on strict position-wise alignment across padded sequences, this length discrepancy causes a semantic mismatch. The model is frequently forced to align semantically disparate linguistic fragments (e.g., matching the root of a verb in the original script with a meaningless subword or phonetic affix in the elongated IPA sequence). This rigid token-level alignment inadvertently introduces positional noise, explaining the limited gains of our fusion strategies and highlighting a fundamental challenge in integrating word-level normalization with subword-level architectures.

\section{Conclusion and Future Work}
This study investigates cross-lingual alignment across four historical stages of Ancient Egyptian. We find that sequence-level multitask learning remains effective under low-resource, high-noise conditions. In particular, translation serves as a strong supervision signal for word-level alignment, outperforming TLM and POS. Normalization combined with KL consistency further improves alignment with English, though it may weaken internal consistency within the four stages of Ancient Egyptian, revealing a trade-off between external comparability and internal coherence.

Aligning historical languages is inherently challenging. The languages differ not only in vocabulary and syntax, but also in scripts and orthographic conventions. These surface differences obscure genealogical links and complicate semantic alignment. Despite these challenges, our results offer a solid foundation for computational research on historical language families and highlight where alignment methods succeed and where they fail. This can support future work on underrepresented and endangered languages with similar structural diversity. 

Looking ahead, several directions remain open. The current normalization scheme can be refined with more linguistically informed rules. Lemma information, especially for pre-Coptic Egyptian, should be incorporated to reduce ambiguity from homographs. Finally, this alignment framework can support future studies of semantic change by tracing cognates across time in a shared embedding space.

\section{Limitations}

\paragraph{Normalization.}
While normalization reduces surface divergence, it introduces several risks:
\begin{itemize}
  \item \textbf{Information loss.} Distinct consonants (e.g., {\egyptology ḥ}, {\egyptology ḫ} and {\egyptology ẖ} $\rightarrow$ \textit{kh}) can be conflated into the same grapheme.
  \item \textbf{False similarity.} Over-aggressive mappings can make unrelated forms appear similar, misleading alignment and evaluation.
  \item \textbf{Oversimplification.} The actual principles governing vowel realization in pre-Coptic Egyptian are more complex than the approximations used in this experiment. 
\end{itemize}

\paragraph{Mismatch between Normalization and Token-Level Alignment.}
As detailed in our error analysis (Section \ref{sec:error_analysis}), integrating word-level normalization with subword-level architectures introduces positional noise due to sequence length divergence. To resolve this structural mismatch, future work should investigate boundary-preserving normalization. Instead of independently tokenizing the normalized text, we could preserve the token boundaries generated from the original Egyptian script and directly map their normalized forms into the vocabulary. Although this risks disrupting context-dependent phonological rules at token boundaries, it guarantees equal input sequence lengths across views, allowing token-level KL divergence and early fusion to operate reliably without requiring complex soft-alignment algorithms like dynamic time warping (DTW).

\paragraph{Granularity of Early Fusion.}
In our experiments, the early embedding fusion strategy consistently underperformed compared to late-stage KL consistency. We hypothesize that a core limitation of our current implementation is the reliance on a single, global learnable scalar ($\alpha$) to control the fusion ratio for all tokens across the entire corpus. While computationally efficient, this global parameter forces a uniform trade-off between original orthography and normalized phonology. Given the heterogeneity across different Egyptian scripts and the varying degrees of normalization opacity, this single-scalar approach lacks flexibility. Future work should investigate fine-grained integration, such as token-level gating mechanisms or language-pair-specific fusion weights, which would allow the model to dynamically balance original and normalized representations on a word-by-word basis.

\paragraph{Lack of lemmatization for pre-Coptic Egyptian.}
We did not incorporate lemma annotations for pre-Coptic Egyptian, omitting a strong source of supervision that could disambiguate homographs and stabilize word-level alignment. This likely contributes to weaker Egyptian-English alignment performance.

\paragraph{Dependence on English Pivot.}

Our current architecture relies heavily on English as a semantic pivot during training. While our intra-Egyptian evaluation demonstrates that the resulting word embeddings do achieve certain direct semantic alignment (as English is entirely absent during this evaluation phase), it remains challenging to strictly isolate how much of this alignment was learned independently of English supervision. Future ablation studies completely removing the English pivot are necessary to quantify direct cross-branch transfer capabilities.

\section{Data and Code Availability}
\label{sec:data_code_availability}

All code, data samples, and evaluation scripts used in this study are available in our GitHub repository (\url{https://github.com/Merythuthor/Egyptian-alignment}). 

Regarding data availability, the resources vary by language branch. For Coptic, we utilize all available Sahidic and Bohairic corpora from Coptic SCRIPTORIUM \citelanguageresource{CopticScriptoriumResource}  (\url{https://github.com/CopticScriptorium/corpora}), which is currently the largest annotated open-access Coptic resource online. 

For pre-Coptic Egyptian, we use a version of the dataset from Thesaurus Linguae Aegyptiae (TLA) for Hieroglyphic and Demotic texts \citelanguageresource{TLA_Hiero, TLA_Demotic}. These data are accessed under research terms and cannot be redistributed in accordance with TLA's policies. However, a version of the dataset sharing the similar annotation format is publicly available at \url{https://huggingface.co/thesaurus-linguae-aegyptiae}, allowing readers to examine the data structure. To ensure maximum reproducibility under these constraints, we provide our comprehensive data processing script (\texttt{data/clean\_corpora.py}) in the GitHub repository of this paper, enabling researchers to recreate our exact experimental setup once they obtain official TLA access.

\section{Acknowledgements}
I would like to sincerely thank Thesaurus Linguae Aegyptiae (TLA) for providing the necessary data access via their official API and Hugging Face and for maintaining such a vital resource for the research community. 

I also extend my sincere thanks to Dr. Caroline T. Schroeder and Dr. Amir Zeldes from the Coptic SCRIPTORIUM for their continuous efforts in maintaining this essential open-access dataset, as well as for their valuable discussions and feedback that greatly benefited the current study. 

I am deeply grateful to my doctoral supervisor, Dr. Friedhelm Hoffmann, for his invaluable instruction in the various historical stages of Ancient Egyptian and for inspiring my interest in language alignment. Furthermore, I sincerely appreciate his continuous support and encouragement in my pursuit of new technologies and interdisciplinary approaches beyond the traditional boundaries of Egyptology.

I would also like to sincerely thank Dr. Daniel A. Werning, Dr. Peter Dils, Dr. Simon D. Schweitzer, Dr. Eliese-Sophia Lincke, Dr. Tonio Sebastian Richter, Dr. Frank Feder, Yihong Liu, Yunting Xie, and Du Cheng for their insightful suggestions and valuable discussions.

\section{Bibliographical References}\label{sec:reference}

\bibliographystyle{lrec2026-natbib}
\bibliography{lrec2026-example}

\section{Language Resource References}
\label{lr:ref}
\bibliographystylelanguageresource{lrec2026-natbib}
\bibliographylanguageresource{languageresource}

\section*{Appendices}
\phantomsection
\subsection*{Appendix A: Sample Sentences across Egyptian Language Stages}
\label{sec:appendix_a}

To provide context for the orthographic diversity of texts within our dataset, we present short illustrative examples of the cleaned text from each of the four language stages used in our training corpus. 

Notice the visual and structural differences: pre-Coptic Egyptian (Hieroglyphic and Demotic) relies on a shared transliteration scheme from the original signs without vowels, whereas the Coptic dialects (Sahidic and Bohairic) use Greek-derived alphabets that also record vowels.

\begin{itemize}
    \item \textbf{Hieroglyphic (Transliteration):} {\egyptology ḏi̯ ⸗s n ⸗k ṯꜣw r fnḏ ⸗k ꜥnḫ ⸗k} \\ 
    \textbf{English Translation:} "She'll give you air to your nose to make you live."

    \item \textbf{Demotic (Transliteration):} {\egyptology ḏd ⸗y n ⸗f r ꞽr wꜥ ṯk ꞽ.ꞽr-ḥr ⸗tn} \\
    \textbf{English Translation:} "I told him to make a letter to you."
    
    \item \textbf{Sahidic (Coptic Script):} {\coptic ϩⲛ ⲡⲉ ϩⲟⲟⲩ ⲉⲛⲧ ⲁ ⲡ ϫⲟⲉⲓⲥ ϣⲁϫⲉ ⲙⲛ ⲙⲱⲩⲥⲏⲥ ϩⲣⲁⲓ ϩⲛ ⲕⲏⲙⲉ} \\
    \textbf{English Translation:} "in the day in which the Lord spoke to Moses in the land of Egypt."
    
    \item \textbf{Bohairic (Coptic Script):} {\coptic ϥ ⲟⲓ ⲛ ϩⲟϯ ⲟⲩⲟϩ ϥ ⲟⲩⲟⲛϩ ⲉⲃⲟⲗ ⲡⲉϥ ϩⲁⲡ ⲉϥⲉ ϣⲱⲡⲓ ⲉⲃⲟⲗ ⲛϧⲏⲧ ϥ} \\
    \textbf{English Translation:} "He is terrible and famous; his judgment shall proceed of himself, and his dignity shall come out of himself."
\end{itemize}

\phantomsection
\subsection*{Appendix B: Tokenization and Multi-Task Input Formats}
\label{sec:appendix_b}

The following example illustrates the multitask input formats and the subword fragmentation caused by normalization. 

\textit{Note: The examples below display the raw internal string representations of the byte-level BPE tokens. As shown, while normalization is performed on the word level during preprocessing, the BPE tokenizer segments the text at the subword (token) level. This inherently leads to severe sequence length mismatch across views (e.g., expanding from 37 tokens to 68 tokens), visually illustrating the structural bottleneck discussed in Section \ref{sec:error_analysis}.}

\begin{flushleft}
\scriptsize
\ttfamily
[Sample: Sahidic Coptic]\\
--- STEP 1 \& 2: Text \& Normalization ---\\
Original : {\coptic ⲱ ⲡ ⲡⲁⲣⲑⲉⲛⲟⲥ ⲉⲧ ⲛϩⲟⲧ ⲁ ⲥ ϣⲱⲡⲉ ⲇⲉ ϩⲙ ⲡⲉ ⲟⲩⲟⲉⲓϣ ⲛ ⲥⲟⲗⲟⲙⲱⲛ}\\
English  : O faithful virgin. Now it came to pass in the time of Solomon\\
LATIN    : oo p parthenos et nhot a s shoope de hm pe ouoejsh n solomoon\\
IPA      : {\ipa oː p partʰənos ət nhot a s ʃoːpə də hm pə uoiʃ n solomoːn}\\
\vspace{1em}
--- STEP 3: Tokens \& Model Inputs ---\\
【Original View】 (Length: 37 tokens)\\
Tokens : ['[CLS]', '\_', '<sah>', '\_â²±', '\_â²¡', '\_â²¡â²ģâ²£â²ĳâ²īâ²Ľâ²Łâ²¥',\\
\ \ \ \ \ \ \ \ \ '\_â²īâ²§', '\_â²ĽÏ©â²Łâ²§', '\_â²ģ', '\_â²¥', '\_Ï£â²±â²¡â²ī', '\_â²ĩâ²ī',\\
\ \ \ \ \ \ \ \ \ '\_Ï©â²Ļ', '\_â²¡â²ī', '\_â²Łâ²©â²Łâ²īâ²ĵÏ£', '\_â²Ľ', '\_â²¥â²Łâ²Ĺâ²Łâ²Ļâ²±â²Ľ',\\
\ \ \ \ \ \ \ \ \ '\_', '[SEP]', '\_', '<eng>', '\_O', '\_faithful', '\_virgin', '.', '\_Now',\\
\ \ \ \ \ \ \ \ \ '\_it', '\_came', '\_to', '\_pass', '\_in', '\_the', '\_time', '\_of',\\
\ \ \ \ \ \ \ \ \ '\_Solomon', '\_', '[SEP]']\\
IDs    : [2, 154, 8, 787, 233, 5949, 379, 23686, 256, 278, 1063, 520,\\
\ \ \ \ \ \ \ \ \ 718, 338, 2916, 208, 11756, 154, 3, 154, 9, 500, 7311, 6282,\\
\ \ \ \ \ \ \ \ \ 23, 5016, 364, 836, 257, 1164, 263, 198, 1097, 215, 11443,\\
\ \ \ \ \ \ \ \ \ 154, 3]\\
\vspace{1em}
【LATIN View】 (Length: 48 tokens)\\
Tokens : ['[CLS]', '\_', '<sah>', '\_o', 'o', '\_p', '\_part', 'hen', 'os', '\_et',\\
\ \ \ \ \ \ \ \ \ '\_nh', 'ot', '\_a', '\_s', '\_sho', 'ope', '\_de', '\_hm', '\_pe', '\_o',\\
\ \ \ \ \ \ \ \ \ 'u', 'oe', 'j', 'sh', '\_n', '\_sol', 'om', 'oon', '\_', '[SEP]', '\_',\\
\ \ \ \ \ \ \ \ \ '<eng>', '\_O', '\_faithful', '\_virgin', '.', '\_Now', '\_it', '\_came',\\
\ \ \ \ \ \ \ \ \ '\_to', '\_pass', '\_in', '\_the', '\_time', '\_of', '\_Solomon', '\_',\\
\ \ \ \ \ \ \ \ \ '[SEP]']\\
IDs    : [2, 154, 8, 202, 87, 223, 1527, 530, 529, 5257, 2597, 295, 203,\\
\ \ \ \ \ \ \ \ \ 204, 17228, 2680, 660, 10372, 758, 202, 93, 2828, 82, 1229,\\
\ \ \ \ \ \ \ \ \ 207, 3167, 288, 3321, 154, 3, 154, 9, 500, 7311, 6282, 23,\\
\ \ \ \ \ \ \ \ \ 5016, 364, 836, 257, 1164, 263, 198, 1097, 215, 11443, 154, 3]\\
\vspace{1em}
【IPA View】 (Length: 68 tokens)\\
Tokens : ['[CLS]', '\_', '<sah>', '\_o', '[UNK]', 'Ĳ', '\_p', '\_p', 'arth',\\
\ \ \ \ \ \ \ \ \ '[UNK]', 'Ļ', 'n', 'os', '\_', '[UNK]', 'Ļ', 't', '\_nh', 'ot',\\
\ \ \ \ \ \ \ \ \ '\_a', '\_s', '\_', 'Ê', 'ĥ', 'o', '[UNK]', 'Ĳ', 'p', '[UNK]', 'Ļ',\\
\ \ \ \ \ \ \ \ \ '\_d', '[UNK]', 'Ļ', '\_hm', '\_p', '[UNK]', 'Ļ', '\_u', 'oi', 'Ê',\\
\ \ \ \ \ \ \ \ \ 'ĥ', '\_n', '\_sol', 'om', 'o', '[UNK]', 'Ĳ', 'n', '\_', '[SEP]',\\
\ \ \ \ \ \ \ \ \ '\_', '<eng>', '\_O', '\_faithful', '\_virgin', '.', '\_Now', '\_it',\\
\ \ \ \ \ \ \ \ \ '\_came', '\_to', '\_pass', '\_in', '\_the', '\_time', '\_of',\\
\ \ \ \ \ \ \ \ \ '\_Solomon', '\_', '[SEP]']\\
IDs    : [2, 154, 8, 202, 1, 171, 223, 223, 4397, 1, 180, 86, 529, 154,\\
\ \ \ \ \ \ \ \ \ 1, 180, 92, 2597, 295, 203, 204, 154, 137, 158, 87, 1, 171,\\
\ \ \ \ \ \ \ \ \ 88, 1, 180, 252, 1, 180, 10372, 223, 1, 180, 354, 4773, 137,\\
\ \ \ \ \ \ \ \ \ 158, 207, 3167, 288, 87, 1, 171, 86, 154, 3, 154, 9, 500,\\
\ \ \ \ \ \ \ \ \ 7311, 6282, 23, 5016, 364, 836, 257, 1164, 263, 198, 1097,\\
\ \ \ \ \ \ \ \ \ 215, 11443, 154, 3]\\
\vspace{1em}
--- STEP 4: Multi-Task Formats ---\\
▶ Task (MLM \& TLM)\\
MLM Input:\\
{[CLS] <sah> }{\coptic ⲱ}{ [MASK] [MASK] }{\coptic ⲉⲧ ⲛϩⲟⲧ ⲁ ⲥ}{ [MASK] }{\coptic ⲇⲉ ϩⲙ ⲡⲉ ⲟⲩⲟⲉⲓϣ ⲛ}{ [MASK] [SEP]}

\vspace{0.5em}
TLM Input:\\
{[CLS] <sah> }{\coptic ⲱ}{ [MASK] [MASK] }{\coptic ⲉⲧ ⲛϩⲟⲧ ⲁ ⲥ}{ [MASK] }{\coptic ⲇⲉ ϩⲙ ⲡⲉ ⲟⲩⲟⲉⲓϣ ⲛ}{ [MASK] [SEP] <eng> [MASK] faithful virgin. [MASK] it [MASK] to pass in the time of Solomon [SEP]}\\
\vspace{1em}
\end{flushleft}

\phantomsection
\subsection*{Appendix C: Examples of Intra-Egyptian Cognate Pairs}
\label{sec:appendix_c}

To evaluate semantic alignment without relying on English, we produced an intra-Egyptian cognate dataset, which is based on TLA's summarization of cognates. Below is a subset of the JSON-formatted data demonstrating the structural format of our evaluation pairs.

\textbf{Dataset Construction Notes:}
\begin{enumerate}
    \item \textbf{Pairwise Mappings:} Cognate pairs are extracted in a strictly pairwise format (e.g., Demotic-to-Sahidic, Sahidic-to-Bohairic). This is because not all historical concepts survived across all four language stages (e.g., Concept \texttt{0014} is only attested in Demotic and Sahidic).
    \item \textbf{Spelling Variants:} A single etymological concept (e.g., Concept \texttt{0005}) often generates numerous paired entries. This occurs because the same word may have multiple attested spelling variants within a single language stage. All possible cross-stage cognate pairs are included to ensure robust evaluation.
\end{enumerate}

\begin{flushleft}
\small
\ttfamily
\{"lang1": "demotic", "word1": "{\egyptology ꜥbyqe}", "lang2": "sahidic", "word2": "{\coptic ⲁϥⲟⲕ}", "source\_concept\_id": "0005"\}\\
\{"lang1": "demotic", "word1": "{\egyptology ꜥbyqe}", "lang2": "sahidic", "word2": "{\coptic ⲁⲃⲱⲕ}", "source\_concept\_id": "0005"\}\\
\{"lang1": "demotic", "word1": "{\egyptology ꜥbq}", "lang2": "sahidic", "word2": "{\coptic ⲁⲃⲟⲟⲕⲉ}", "source\_concept\_id": "0005"\}\\
\{"lang1": "demotic", "word1": "{\egyptology ꜥbqy}", "lang2": "sahidic", "word2": "{\coptic ⲁⲃⲟⲕ}", "source\_concept\_id": "0005"\}\\
\{"lang1": "demotic", "word1": "{\egyptology ꜥbyqe}", "lang2": "bohairic", "word2": "{\coptic ⲁⲃⲟⲕⲓ}", "source\_concept\_id": "0005"\}\\
\{"lang1": "demotic", "word1": "{\egyptology ꜥbq}", "lang2": "bohairic", "word2": "{\coptic ⲁⲃⲟⲕ}", "source\_concept\_id": "0005"\}\\
\{"lang1": "sahidic", "word1": "{\coptic ⲁϥⲟⲕ}", "lang2": "bohairic", "word2": "{\coptic ⲁⲃⲟⲕⲓ}", "source\_concept\_id": "0005"\}\\
\{"lang1": "sahidic", "word1": "{\coptic ⲁⲃⲱⲕ}", "lang2": "bohairic", "word2": "{\coptic ⲁⲃⲟⲕ}", "source\_concept\_id": "0005"\}\\
\{"lang1": "hieroglyphic", "word1": "{\egyptology ꜥqy}", "lang2": "demotic", "word2": "{\egyptology ꜥyq}", "source\_concept\_id": "0010"\}\\
\{"lang1": "hieroglyphic", "word1": "{\egyptology ꜥqy}", "lang2": "sahidic", "word2": "{\coptic ⲁⲉⲓⲕ}", "source\_concept\_id": "0010"\}\\
\{"lang1": "demotic", "word1": "{\egyptology ꜥlwꜣ}", "lang2": "sahidic", "word2": "{\coptic ⲁⲗ}", "source\_concept\_id": "0014"\}
\end{flushleft}

\phantomsection
\subsection*{Appendix D: Evaluation Dataset Statistics}
\label{sec:appendix_d}

Table~\ref{tab:eval_stats} details the number of valid word pairs ($N$) used for the semantic alignment evaluation across all language pairs and conditions. A word pair is considered valid if both constituent words appear in the held-out context pool and possess contextualized embeddings for cosine similarity computation. The numbers reflect the final filtered and deduplicated evaluation sets used in all reported experiments.
\begin{table}[H]
\centering
\resizebox{\columnwidth}{!}{%
\begin{tabular}{lllc}
\toprule
\textbf{Evaluation Task} & \textbf{Category} & \textbf{Language Pair} & \textbf{Valid Pairs ($N$)} \\
\midrule
\multirow{8}{*}{Intra-Egyptian} & \multirow{4}{*}{Cross-Branch (Heterograph)} & Bohairic--Demotic & 396 \\
& & Bohairic--Hieroglyphic & 294 \\
& & Demotic--Sahidic & 708 \\
& & Hieroglyphic--Sahidic & 644 \\
\cmidrule{2-4}
& \multirow{2}{*}{Within-Branch (Heterograph)} & Bohairic--Sahidic & 318 \\
& & Demotic--Hieroglyphic & 318 \\
\cmidrule{2-4}
& \multirow{2}{*}{Within-Branch (Homograph)} & Bohairic--Sahidic & 82 \\
& & Demotic--Hieroglyphic & 181 \\
\midrule
\multirow{8}{*}{Egyptian--English} & \multirow{2}{*}{Hieroglyphic--English} & Seen & 6,883 \\
& & Unseen & 1,825 \\
\cmidrule{2-4}
& \multirow{2}{*}{Demotic--English} & Seen & 1,778 \\
& & Unseen & 468 \\
\cmidrule{2-4}
& \multirow{2}{*}{Sahidic--English} & Seen & 1,967 \\
& & Unseen & 355 \\
\cmidrule{2-4}
& \multirow{2}{*}{Bohairic--English} & Seen & 1,598 \\
& & Unseen & 354 \\
\bottomrule
\end{tabular}%
}
\caption{Number of valid word pairs ($N$) utilized for each specific evaluation group.}
\label{tab:eval_stats}
\end{table}

\end{document}